\documentclass[letterpaper, 10 pt, conference]{ieeeconf}  
\usepackage{cite}
\usepackage{amsmath,amssymb}
\usepackage{tikz}
\usetikzlibrary{positioning, calc, fit, matrix, arrows.meta}
\usepackage{graphicx}
\usepackage{subcaption}
\usepackage{booktabs}
\usepackage{array}
\usepackage{multirow}
\usepackage{url}
\usepackage[hidelinks,breaklinks=true]{hyperref}
\newtheorem{assumption}{Assumption}
\newtheorem{definition}{Definition}
\newcounter{appsec}
\renewcommand{\theappsec}{\Alph{appsec}}
\newcommand{\appsection}[1]{%
  \refstepcounter{appsec}%
  \noindent\textbf{\theappsec\quad #1}\par\nobreak\medskip}

\IEEEoverridecommandlockouts                              

\title{\LARGE \bf
Anomaly-Informed Confidence Calibration \\for Vision-Based Safety Prediction
}

\author{
Zhenjiang Mao$^{1\dagger}$, Jiawen Wu$^{1\dagger}$, Gabriel Wagner$^{1}$, Zhongzheng Zhang$^{1}$  and Ivan Ruchkin$^{1}$%
\thanks{$^{\dagger}$First co-authors: equal contribution.}%
\thanks{$^{1}$Trustworthy Engineered Autonomy (TEA) Lab,
Department of Electrical and Computer Engineering,
University of Florida, Gainesville, FL 32611, USA
{\tt\small \{z.mao, jiawenwu, gabriel.wagner, i.ruchkin\}@ufl.edu}}%
\thanks{
This work was supported in part by the NSF Grant CNS 2513076. Any opinions, findings, or conclusions expressed in this material are those of the authors and do not necessarily reflect the views of the National Science Foundation (NSF) or the U.S. Government.
The authors would like to thank Arion Stern, Liam Cade McGlothlin, Josh Kumar, Jay Nibhanupudy, and Chengyu Li for collecting and analyzing DonkeyCar data and for investigating related ideas.
}
}

\begin{document}

\maketitle
\thispagestyle{empty}
\pagestyle{empty}

\begin{abstract}
Reliable confidence estimates are important for safely deploying vision-based controllers in autonomous racing, where safety predictions must be derived from camera images, yet modern predictors become dangerously overconfident under test-time distribution shifts. A commonly assumed \emph{perception-dynamics gap} holds that observation-space anomaly signals, such as autoencoder reconstruction error, should miss dynamics anomalies (e.g., actuation bias, latency) that originate outside the image. We interrogate this gap with an \textit{Anomaly-Informed Test-Time Calibration} approach that, without retraining any model component, fuses a perceptual score (reconstruction error) and a disentangled dynamics score (epistemic uncertainty and control-stream statistics) from a world model to condition a lightweight temperature-scaling calibrator, aided by test-time augmentation. On a physical DonkeyCar under four real-world anomaly protocols unseen during training (darkness, blur, actuation bias, processing latency), it cuts average expected calibration error from 0.184 to 0.116, a 37\% improvement over the best baseline, without modifying the base predictor. Our central finding is that in a tightly coupled closed loop, the perception-dynamics gap is not prominent. Specifically, a dynamics fault degrades the trajectory until the observation is itself out-of-distribution, so the perceptual score alone already captures most dynamics anomalies and accounts for the calibration gains. As a result, the disentangled dynamics score adds little further calibration but remains a direct, interpretable signal that aids out-of-distribution detection of dynamics and control faults. \\
Video summary: \url{https://youtu.be/TWemi_OTZQI}
\end{abstract}

\section{INTRODUCTION}

Reliable confidence estimates are key to deploying vision-based robotic policies in safety-critical settings, yet modern predictors often exhibit miscalibration and dangerously overconfident behavior under test-time anomalies~\cite{lee2018simple}. The safety evaluator must therefore provide confidence scores that reflect the true likelihood of remaining safe~\cite{mao2024howsafe,mao2025howsafewill}. This confidence miscalibration is significantly exacerbated when encountering distribution shifts during deployment~\cite{guo2017calibration,ovadia2019can}. A key obstacle is what we call the perception-dynamics gap: standard anomaly scores operate entirely in observation space, so a failure that originates in the dynamics --- actuator bias, control latency, or unmodeled friction --- reaches them only \emph{indirectly} and late, through the image degradation it eventually induces, rather than being flagged directly from the control stream as it arises.


Standard post-hoc calibration methods, such as temperature scaling, rely on a single learned scalar and lose performance substantially under dataset shift~\cite{tomani2021posthoc}. Richer uncertainty quantification techniques (Monte Carlo (MC) Dropout~\cite{gal2016dropout}, Deep Ensembles~\cite{lakshminarayanan2017simple}) improve robustness but incur computational costs that conflict with real-time control, and treat all anomalies as a monolithic source of uncertainty without distinguishing perceptual from dynamics-driven deviations~\cite{tian2021exploring,yu2022robust}.

\begin{figure*}[t]
  \centering
  \includegraphics[width=\textwidth]{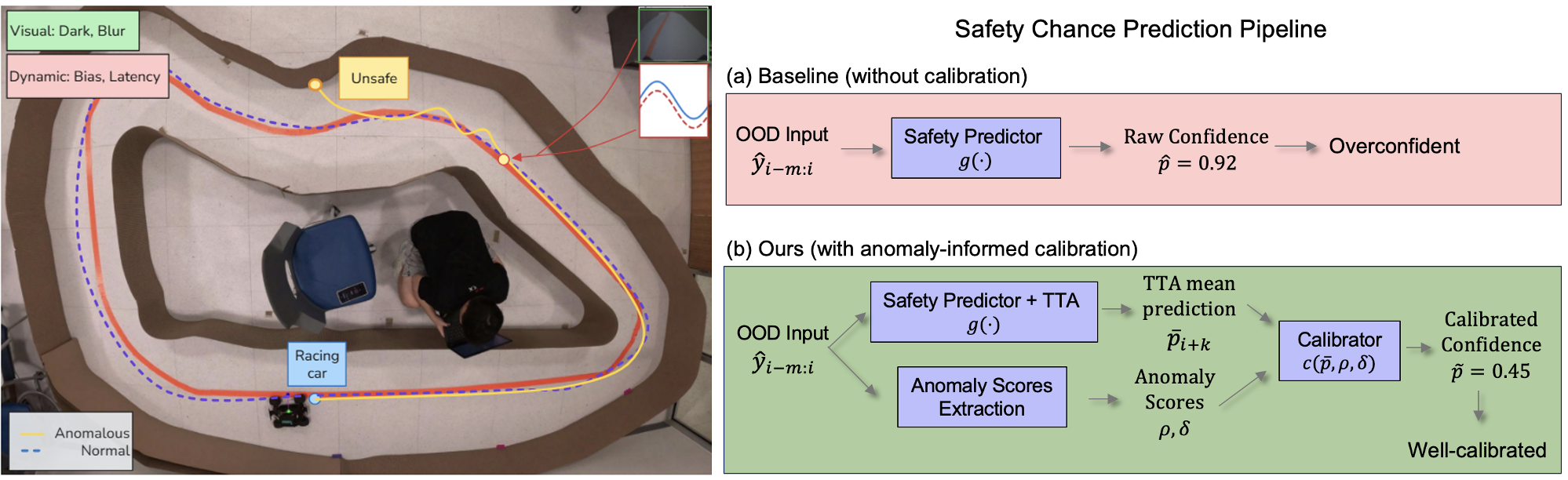}
  \caption{Problem overview on the DonkeyCar platform.
    \textbf{Left:} Two distribution shifts encountered at deployment: \emph{visual}
    anomalies (e.g., darkness, blur) that corrupt the observation $y_t\!=\!o(x_t)$, and
    \emph{dynamics} anomalies (e.g., steering bias, latency) that corrupt the
    transition $x_{t+1}\!=\!f(x_t,u_t)$.
    Under normal conditions, the car stays on-track (blue dashes); under out-of-distribution (OOD)
    shifts, the trajectory degrades and leaves the track (yellow solid).
    \textbf{Right:} Our approach fuses TTA-enhanced predictions with anomaly
    scores $\rho$ and $\delta$ via a learned calibrator, producing reliable
    confidence suitable for downstream decision-making.}
  \label{fig:problem-overview}
  \vspace{-3mm}
\end{figure*}

\looseness=-1
Distribution shifts can corrupt the observation model (visual noise, lighting changes) or the dynamical model (actuation noise, unmodeled physics)~\cite{ha2018world,hendrycks2019robustness}, as illustrated on the left of Fig.~\ref{fig:problem-overview}. 
Variational Autoencoder (VAE) reconstruction error~\cite{an2015variational}, a popular anomaly signal, quantifies per-frame deviation from the learned visual manifold, so it flags dynamics anomalies (e.g., actuation lag, frame-freezing) only once the diverging latent trajectory has corrupted the images, not while individual frames still look plausible.
This motivates a calibration strategy conditioned on \emph{disentangled} anomaly evidence spanning both perception and dynamics.

Our core idea is to let the calibrator observe \emph{the kind} of anomaly and \emph{how severe} it is, rather than relying on a single fixed temperature. Thus, we propose a real-time \textit{anomaly-informed test-time calibration} framework for vision-based robotic safety prediction. Our method repurposes signals from an anomaly-unaware world model to construct two complementary anomaly scores: a perceptual score from VAE reconstruction error, and a dynamics score derived from epistemic uncertainty estimates and control-stream statistics. 
We additionally apply lightweight \textit{test-time augmentation (TTA)} to smooth predictions under visual shift.
These scores feed a lightweight temperature-scaling module that adjusts the effective temperature, reducing confidence under anomalies while preserving nominal behavior. Note that we treat this disentanglement as a hypothesis to test, not a foregone benefit --- and find it largely does not pay off in this closed loop (Section~\ref{sec:results}).

The primary contributions of this work are as follows:
\begin{enumerate}
\item A \emph{test-time calibration framework} that fuses anomaly evidence from a world model with test-time augmentation to adjust safety confidence without retraining.
\item An empirical finding that the perception-dynamics gap is not significant in a tightly coupled closed loop: since dynamics faults propagate into the observation stream, a disentangled dynamics score adds little calibration over the perceptual score alone, though it remains a direct, interpretable channel that aids OOD detection on pure actuation faults.
\item A \emph{physical DonkeyCar validation} showing that calibration trained on synthetic augmentations transfers to qualitatively different real-world corruptions, with a 37\% reduction in calibration error over the best baseline across four unseen OOD protocols.
\end{enumerate}

The remainder of this paper is organized as follows. Section~II reviews related work. Section~III establishes the formal problem setting. Section~IV details our anomaly-informed calibration framework. Section~V describes the experimental setup on a physical DonkeyCar platform and analyzes the results under four real-world distribution shift protocols. Section~VI concludes with a discussion of limitations and future directions.

\section{RELATED WORK}
\label{sec:related_work}

Confidence calibration aligns a model's prediction confidence with its empirical accuracy. Post-hoc calibrators range from parametric (beta calibration, Bayesian Binning) to density-aware approaches~\cite{tomani2023beyond}; the latter, modulating temperature via $k$-nearest-neighbor (KNN) feature density, is most related to ours.
The primary challenge static methods face is degradation under drift~\cite{tomani2021posthoc,ovadia2019can}. Approaches such as multi-domain temperature scaling~\cite{yu2022robust} address this by explicitly leveraging heterogeneity across multiple training domains, MC Dropout~\cite{gal2016dropout} and Deep Ensembles~\cite{lakshminarayanan2017simple} through multiple passes, and test-time augmentation through stochastic input averaging~\cite{hekler2023tta}.
Our method differs by instead conditioning temperature on disentangled anomaly scores while keeping the base predictor frozen.

Anomaly detection identifies anomalous data by observing OOD indicators. Standard OOD indicators include the maximum softmax probability (MSP)~\cite{hendrycks2017baseline}, Mahalanobis distance in feature space~\cite{lee2018simple}, feature norms~\cite{yu2023featnorm} and energy scores. For visual anomaly detection, techniques such as attention-map monitoring, vision-language semantic detection~\cite{mao2024language}, KNN feature distance, activation truncation~\cite{sun2021react}, and VAE reconstruction error~\cite{an2015variational} are used.
Outlier exposure~\cite{hendrycks2019oe} trains against auxiliary outlier data, with later work synthesizing virtual outliers in feature space~\cite{du2022vos}. The primary challenges anomaly detection and confidence calibration face are covariate and concept shift~\cite{tian2021exploring}. Covariate shift reflects style-level changes (e.g., noise) and concept shift reflects semantic-level changes. Each requires distinct score functions for reliable OOD detection and confidence calibration.
Unlike detection methods that produce binary flags, we use continuous anomaly scores to quantitatively adjust confidence and apply outlier exposure to train a separate calibrator rather than the detector itself.
Visual detection methods, like our proposed perception score $\rho$, detect anomalies in observation. We additionally propose a dynamics score~$\delta$ to detect anomalies based on the internal dynamics model and control stream.
A complementary line of work acts on the observation itself, repairing corrupted images before they reach the controller so that the downstream policy sees nominal inputs~\cite{sobolewski2025imagerepair}. We instead leave both the input and the base predictor untouched and adjust only the reported confidence, which avoids the risk of a repair step hallucinating structure that was never observed and keeps nominal behavior unchanged when no anomaly is present.

The closest antecedents to our work are world-model-based safety prediction~\cite{mao2024howsafe,mao2025howsafewill,mao2024zeroshot}, and autoencoder misbehavior detection~\cite{stocco2020misbehaviour}.
The world models used in this work follow the VAE-RNN world model paradigm~\cite{ha2018world}, which compresses high-dimensional observations into a low-dimensional latent representation and models temporal dynamics over that latent space. The resulting prediction is used by a compact controller.
This architecture was subsequently scaled by the PlaNet and Dreamer line of work~\cite{hafner2019planet,hafner2020dream}, which pairs a deterministic recurrent state with a stochastic latent in a recurrent state-space model and learns behavior from latent imagination. DreamerV3 used this approach to master diverse control tasks with a single configuration~\cite{hafner2025mastering}.
Recent efforts pursue \emph{physically interpretable} world models whose latent factors map to meaningful physical quantities~\cite{peper2025principles,mao2026physically}.
Our study instantiates a similar convolutional VAE encoder/decoder and a ConvLSTM latent predictor. We repurpose its internal reconstruction error and latent-rollout uncertainty as perceptual and dynamics anomaly scores. These scores drive calibration, while keeping the world model frozen.
Combined with the end-to-end driving approach~\cite{bojarski2016end} of the DonkeyCar robotic platform, we contribute test-time uncertainty quantification to an otherwise anomaly-unaware world model.

\section{PRELIMINARIES}
\label{sec:preliminaries}

\looseness=-1

We formalize the vision-based safety prediction problem and then define the anomaly types and calibration objective.


\begin{definition}[Vision-based Robotic System]
A vision-based system $s = (\mathcal{X}, \mathcal{Y}, \mathcal{U}, x_0, h, f, o)$ consists of a continuous state space $\mathcal{X}$, an image observation space $\mathcal{Y}$, a control action space $\mathcal{U}$, an initial state $x_0 \sim P_0$, a vision-based controller $h: \mathcal{Y} \to \mathcal{U}$, an unknown dynamical model $f: \mathcal{X} \times \mathcal{U} \to \mathcal{X}$ with $x_{t+1} = f(x_t, u_t)$, and an unknown observation model $o: \mathcal{X} \to \mathcal{Y}$ with $y_t = o(x_t)$.
\end{definition}

\looseness=-1

We instantiate this on an autonomous racing platform where $\mathcal{X}$ is vehicle pose and velocity, $\mathcal{Y}$ is camera images, and $\mathcal{U}$ is steering and throttle.
We denote the closed-loop trajectory $\tau_t = (x_t, u_t, x_{t+1}, u_{t+1}, \ldots)$ under $u_t = h(o(x_t))$ and binary safety indicator $\psi(x): \mathcal{X} \to \{0,1\}$.

\begin{definition}[Safety Confidence Predictor]
\label{def:predictor}
Given history length $n$ and prediction horizon $k$, a \emph{safety confidence predictor} $g$ takes the last $n$ image observations $y_{i-n+1:i}$ as input and outputs a predicted safety confidence $\hat{p}_{i+k} \in [0,1]$, representing
$\hat{p}_{i+k} \approx P(\psi(x_{i+k})=1 \mid y_{i-n+1:i}, h). $
The predicted safety label is obtained by thresholding: $\hat{s}_{i+k} := \mathbf{1}[\hat{p}_{i+k} \geq 0.5]$.
\end{definition}
In our implementation, $n=1$, so $g$ takes a single image. Because the controller is fixed, the predicted confidence implicitly accounts for the future actions the closed-loop policy will take.


\noindent
\textbf{Anomalies.}
 We categorize these shifts into two classes: visual anomalies and dynamic anomalies. 
\emph{Visual anomalies} change observation function $o$ into some alternative $o' \neq o$. This change directly perturbs the image stream $\{y_t\}$ without altering the underlying dynamics. Examples include sensor degradation, lighting changes, lens blur, and visual occlusions.
\looseness=-1
\emph{Dynamics anomalies} change the dynamics function $f$ into some alternative $f' \neq f$, inducing deviations in the state trajectory $\{x_t\}$. Examples include actuation bias, control latency, and changes in surface friction and payload.

To account for such shifts, we assume access to anomaly features that summarize the severity of each type of distributional deviation. The anomaly injection protocols used for evaluation are described in Section~\ref{sec:experimental_setup}; the full dynamics feature definitions are provided in Appendix~\ref{app:dynamics_features}.


\begin{definition}[Anomaly Features]
At each time $i$, we assume access to an \emph{anomaly feature vector} $\mathcal{E}_i = (\rho_i, \delta_i)$:
\begin{itemize}
    \item $\rho_i \in [0,1]$ quantifies the severity of visual distribution shift (perception anomaly),
    \item $\delta_i \in [0,1]$ quantifies the severity of dynamics distribution shift (dynamics anomaly).
\end{itemize}
\end{definition}

The safety prediction problem thus decomposes into three sub-tasks: (1) \textit{base safety prediction} --- estimating raw safety confidence $\hat{p}_{i+k}$ without anomalies; (2) \emph{anomaly quantification} --- estimating the anomaly features $\mathcal{E}_i$ that quantify the severity of distribution shift; and (3) \emph{confidence calibration} --- mapping the raw confidence $\hat{p}_{i+k}$ and the anomaly features $\mathcal{E}_i$ to a better-calibrated confidence
 $\tilde{p}_{i+k}$. 
\begin{definition}[Anomaly-Informed Calibration] \label{def:AIC}
An \emph{anomaly-informed calibration function} $\mathcal{C}: [0,1] \times \mathcal{E} \rightarrow [0,1]$ maps a raw safety confidence $\hat{p}_{i+k}$ and anomaly features $\mathcal{E}_i = (\rho_i, \delta_i)$ to a calibrated confidence
$\tilde{p}_{i+k} = \mathcal{C}(\hat{p}_{i+k}, \mathcal{E}_i)$ such that $\tilde{p}_{i+k}$ better reflects the true safety probability $P(\psi(x_{i+k})=1 \mid y_{i-n+1:i}, h)$.
\end{definition}

In practice, the specific corruptions encountered at test time are unknown a priori. To train the calibrator, we use generic augmentations as a structural proxy for real-world corruptions and unmodeled dynamics, enabling the model to learn the relationship between anomaly intensity and predictor reliability.


\begin{definition}[Data Regimes]
We consider three datasets: an in-distribution (ID) set $\mathcal{D}_{\mathrm{in}}$ (split into $\mathcal{D}_{\mathrm{in}}^{\mathrm{train}}$, $\mathcal{D}_{\mathrm{in}}^{\mathrm{cal}}$, and an ID test set $\mathcal{D}_{\mathrm{in}}^{\mathrm{test}}$), a synthetically augmented set $\mathcal{D}_{\mathrm{aug}}$, and a test-time anomalous set $\mathcal{D}_{\mathrm{ood}}$ subject to real-world shifts. The calibrator is trained on $\mathcal{D}_{\mathrm{in}}^{\mathrm{cal}} \cup \mathcal{D}_{\mathrm{aug}}$; $\mathcal{D}_{\mathrm{in}}^{\mathrm{test}}$ and $\mathcal{D}_{\mathrm{ood}}$ are reserved for evaluation only.
\end{definition}

For the calibrator to generalize from synthetic augmentations to unseen real-world anomalies, we require a structural relationship in the anomaly feature space:


\begin{assumption}[Augmentation-Anomaly Transferability]
\label{assump:transfer}
Let $c$ denote an arbitrary corruption (augmentation or real-world anomaly) applied with severity level $s \geq 0$, and let $\mathcal{E}_i(c, s) = (\rho_i(c,s),\, \delta_i(c,s))$ denote the resulting anomaly features. We assume:
\begin{enumerate}
    \item \textbf{Monotonicity:} For every corruption type $c$, the anomaly features are non-decreasing in severity: $s_1 \leq s_2 \implies \rho_i(c, s_1) \leq \rho_i(c, s_2)$ and $\delta_i(c, s_1) \leq \delta_i(c, s_2)$.
    \item \textbf{Approximate Sufficiency:} The true safety probability conditioned on the anomaly features is approximately invariant to the corruption type, i.e.,
    \begin{equation}
    \begin{aligned}
    \Pr\!\big(\psi(x_{i+k})=1 \mid \hat{p}_{i+k},\, \mathcal{E}_i,\, c\big)
    \\ \approx\;
    \Pr\!\big(\psi(x_{i+k})=1 \mid \hat{p}_{i+k},\, \mathcal{E}_i\big).
    \end{aligned}
    \end{equation}
\end{enumerate}
\end{assumption}

\medskip
Given the above formulation, we now describe the metrics used to evaluate calibration quality.
We measure calibration quality with Expected Calibration Error (ECE)~\cite{guo2017calibration} and discriminative ability with the Area Under the ROC Curve (AUROC)~\cite{fawcett2006roc}. 


\section{ANOMALY-INFORMED CONFIDENCE}
\label{sec:approach}

\begin{figure*}[t]
  \centering
  \includegraphics[width=0.8\textwidth]{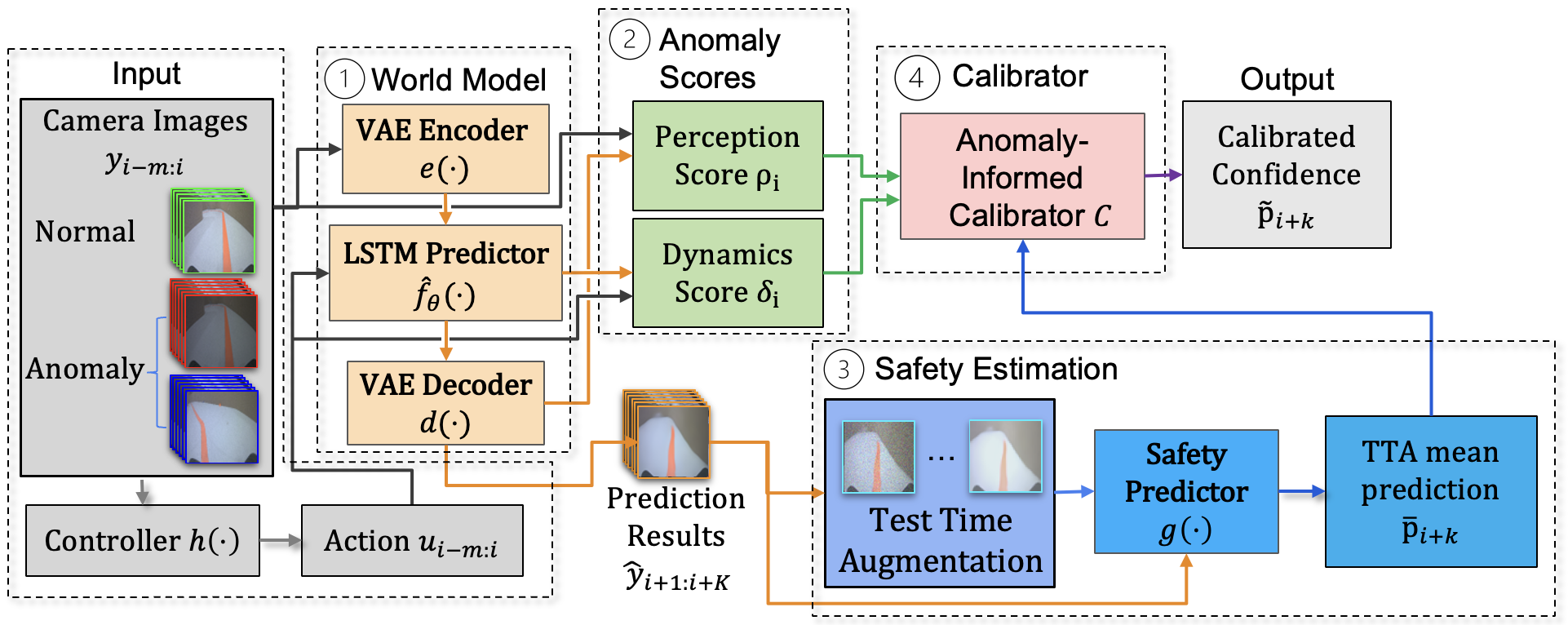}
    \caption{Approach overview. At each time step~$i$, the world model extracts a perception score~$\rho_i$ and a dynamics score~$\delta_i$ from its internal inference errors. The safety predictor~$g$ is evaluated under test-time augmentation to produce a TTA-averaged prediction~$\bar{p}_{i+k}$, which is then calibrated via anomaly-conditioned temperature scaling to yield the final confidence~$\tilde{p}_{i+k}$.}
  \label{fig:approach_overview}
  \vspace{-3mm}
\end{figure*}

The pipeline (Fig.~\ref{fig:approach_overview}) has four stages:
(1)~the \emph{world model} ($e(\cdot)$, $d(\cdot)$, $\hat{f}_\theta(\cdot)$) encodes the camera history $y_{i-m_w+1:i}$ and rolls out the latent dynamics, exposing its internal reconstruction and prediction errors for scoring;
(2)~the encoder branch yields the perception score $\rho_i$ from reconstruction error and the predictor branch yields the dynamics score $\delta_i$ from rollout uncertainty;
(3)~the \emph{safety predictor} $g(\cdot)$ is evaluated on the current observation under $M$ test-time augmentations and averaged into the TTA mean prediction $\bar{p}_{i+k}$;
(4)~the \emph{anomaly-informed calibrator} $C$ applies anomaly-conditioned temperature scaling to $\bar{p}_{i+k}$, conditioned on $\rho_i$ and $\delta_i$, to produce the calibrated confidence $\tilde{p}_{i+k}$.
Stages~(1)--(3) run online; stage~(4) uses temperature parameters fit offline.

\noindent
\textbf{Anomaly Score Extraction.}
Rather than treating anomaly detection as a binary decision, we extract scalar scores from the world model's internal errors, providing the calibrator with graded evidence of distribution shift.

\begin{definition}[Learned World Model]
\label{def:world_model}
A \emph{learned world model} $\omega = (e, d, \hat{f}_\theta)$ consists of an encoder $e: \mathcal{Y} \to \mathcal{Z}$, a decoder $d: \mathcal{Z} \to \mathcal{Y}$, and latent dynamics $\hat{f}_\theta: \mathcal{Z} \times \mathcal{U} \to \mathcal{Z}$, where $\mathcal{Z}$ is a convolutional latent space ($z_t \in \mathbb{R}^{64\times4\times4}$).
\end{definition}
\noindent The world model is trained on $\mathcal{D}_{\text{in}}$ and remains frozen at test time. At each observation time $i$, we compute $\mathcal{E}_i = (\rho_i, \delta_i)$:

\subsubsection{Perception Impact Score~\texorpdfstring{$\rho_i$}{rho\_i}}
We quantify visual anomaly severity as the mean per-pixel reconstruction error of~$d$ over the last $m_w$ context frames:
\begin{equation}
    \rho_i^{\mathrm{raw}} \;=\; \frac{1}{m_w}\sum_{j=i-m_w+1}^{i}
      \frac{1}{HWC}\bigl\lVert y_j - d(e(y_j))\bigr\rVert_F^2\,,
    \label{eq:rho-raw}
\end{equation}
where $H,W,C$ are image dimensions and $\frac{1}{HWC}\lVert\cdot\rVert_F^2$ is the per-pixel MSE.

We average over $m_w$ frames to suppress frame-level noise in reconstruction error (e.g., auto-exposure jitter); $\rho_i$ thus captures sustained perceptual shift rather than transient spikes.
The raw score is then mapped to $[0,1]$ using in-distribution statistics:
\begin{equation}
    \rho_i \;=\; \mathrm{clip}_{[0,1]}\!\left(
      \frac{\rho_i^{\mathrm{raw}} - \mu_\rho}{\alpha\,\sigma_\rho}
    \right),\quad \alpha=2,
    \label{eq:rho}
\end{equation}
\looseness=-1
where $\mu_\rho$ and $\sigma_\rho$ are the mean and standard deviation of $\rho^{\mathrm{raw}}$ estimated on the held-out calibration split $\mathcal{D}_{\mathrm{in}}^{\mathrm{cal}}$.
The scale $\alpha{=}2$ maps $\mu_\rho+2\sigma_\rho$ to $1$; values below $\mu_\rho$ are clipped to $0$.


\subsubsection{Dynamics Impact Score~\texorpdfstring{$\delta_i$}{delta\_i}}
The dynamics impact score captures anomalies in the state-transition space that may leave the observation visually plausible while compromising the latent trajectory.
For each context window, we form a 12-dimensional vector $\mathbf{f}_i^{K_{\max}}$ of generic, anomaly-agnostic descriptors derived from the approximate dynamics model~$\hat{f}_\theta$ and the control stream, grouped as:
\begin{itemize}
    \item \textbf{MC Dropout uncertainty (3):} rollout-mean, rollout-max, and volatility of per-step latent std estimated via MC Dropout on~$\hat{f}_\theta$ over a $K_{\max}$-step rollout;
    \item \textbf{Action patterns (6):} the longest steering hold (overall and during active turns), the fraction of active-turn steps that are held, the steering reversal rate, jerk, and mean steering from $u_{i-m_w+1:i}$;
    \item \textbf{Temporal correlations (3):} the steering-latent cross-correlation peak, the high-frequency spectral ratio, and the steering-latent response lag, computed within the context window.
\end{itemize}
\looseness=-1
All features are computed causally from the available context $(y_{i-m_w+1:i},u_{i-m_w+1:i})$ and stochastic rollouts of~$\hat{f}_\theta$ --- no future observations are required. Therefore, $\delta_i$ is available test-time at time~$i$.
See Appendix~\ref{app:dynamics_features} for full feature definitions.

We compute the dynamics score as the Mahalanobis distance~\cite{lee2018simple} from the in-distribution feature manifold, normalized to $[0,1]$ using the same z-score clipping as for the perception score (Eq.~\ref{eq:rho}):
\begin{equation}
    \delta_i \;=\; \mathrm{clip}_{[0,1]}\!\left(
      \frac{d_{\text{Maha}}(\mathbf{f}_i^{K_{\max}},\; \mu_{\text{in}},\; \Sigma_{\text{in}}) - \mu_\delta}{\alpha\,\sigma_\delta}
    \right),\quad \alpha=2,
    \label{eq:delta}
\end{equation}
\looseness=-1
where $\mu_{\text{in}}$ and $\Sigma_{\text{in}}$ are the sample mean and covariance of
$\{\mathbf{f}_i^{K_{\max}}\}_{i \in \mathcal{D}_{\mathrm{in}}^{\mathrm{cal}}}$, and $\mu_\delta$, $\sigma_\delta$ are the mean and standard deviation of the Mahalanobis distance on $\mathcal{D}_{\mathrm{in}}^{\mathrm{cal}}$.
We apply covariance regularization ($\Sigma_{\text{in}} \leftarrow \Sigma_{\text{in}} + \lambda I$) for numerical stability.

\begin{figure}[t]
  \centering
  \includegraphics[width=\linewidth]{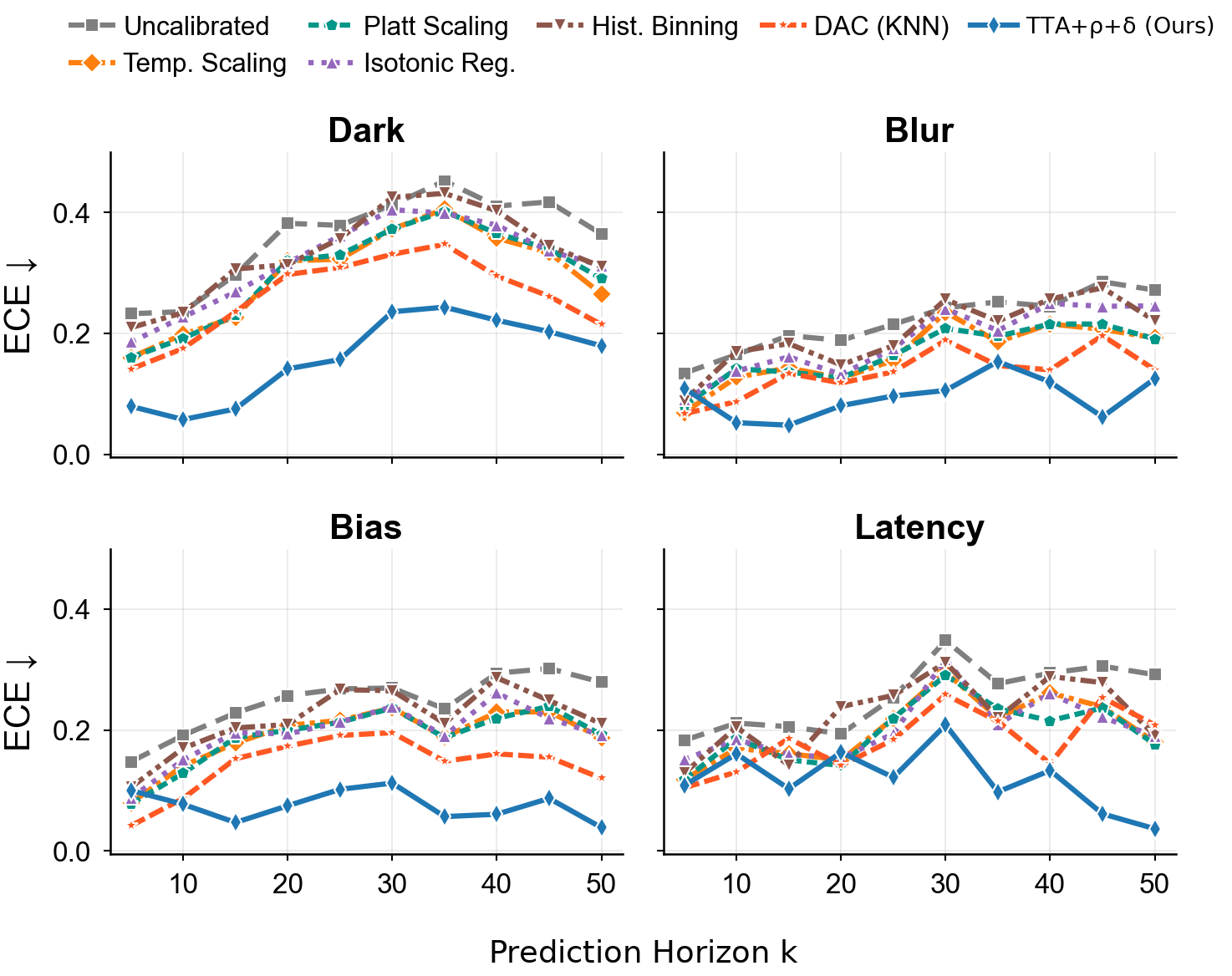}
  \caption{Baseline comparison: ECE ($\downarrow$) vs.\ prediction horizon~$k$ under four OOD protocols. Our method (TTA+$\rho$+$\delta$) consistently achieves the lowest calibration error.}
  \label{fig:baseline_ece}
  \vspace{-7mm}
\end{figure}

\begin{figure}[h]
  \centering
  \includegraphics[width=\linewidth]{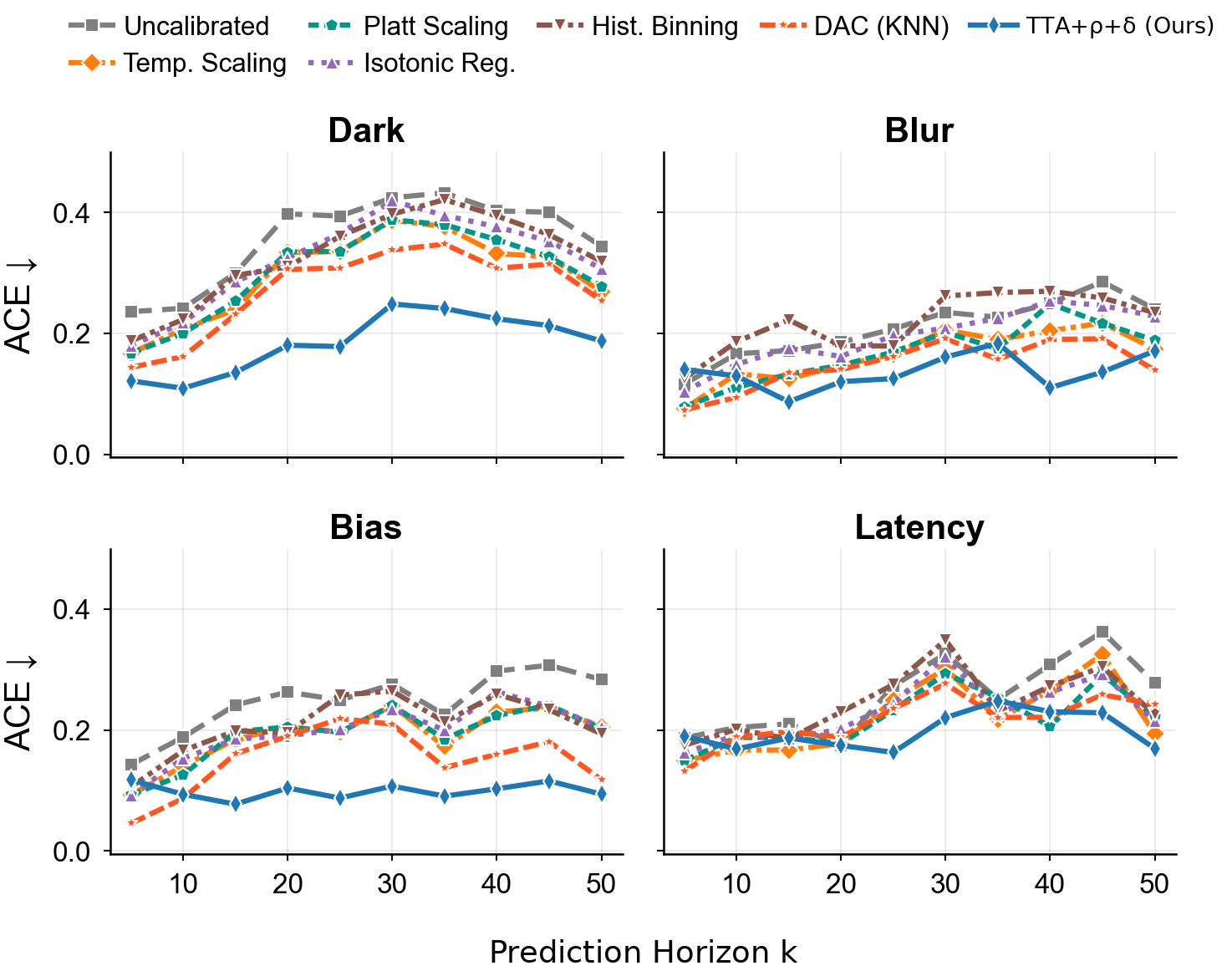}\\[4pt]
  \includegraphics[width=\linewidth]{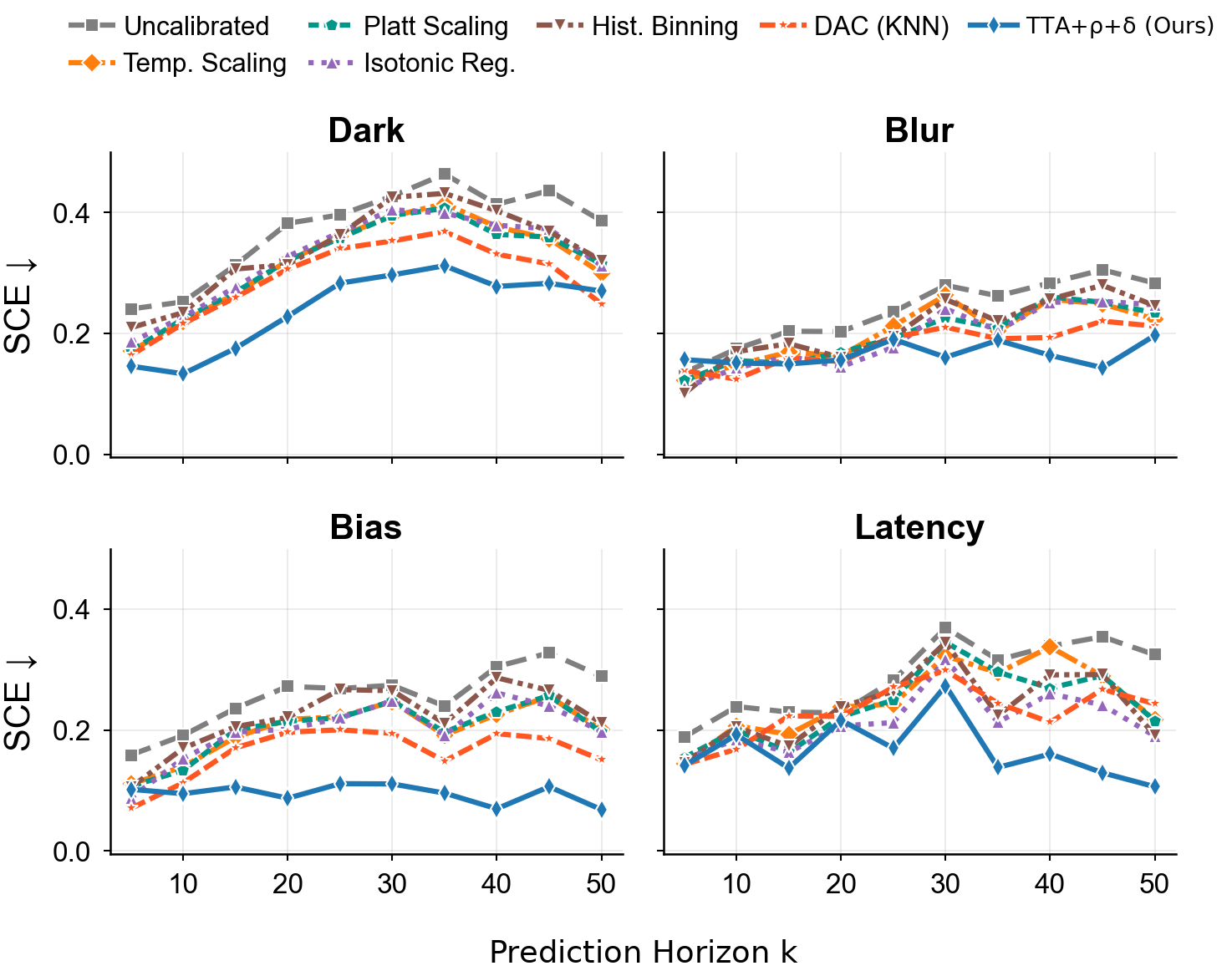}
  \caption{Alternative calibration metrics vs.\ prediction horizon~$k$. \textbf{Top:} ACE ($\downarrow$). \textbf{Bottom:} SCE ($\downarrow$). Method rankings are consistent with the ECE results in Fig.~\ref{fig:baseline_ece}.}
  \label{fig:ace_sce}
  \vspace{-3mm}
\end{figure}


\noindent
\textbf{Test-Time Calibration.} Given the raw confidence $\hat{p}_{i+k}$ from Stage~1 and the anomaly scores $(\rho_i, \delta_i)$ from Stage~2, the calibration stage produces a calibrated probability through two mechanisms: test-time augmentation (Stage~3) and anomaly-conditioned temperature scaling (Stage~4).
\subsubsection{Visual Robustness via Test-Time Augmentation}
\label{sec:tta}

Following~\cite{hekler2023tta}, let $\{\mathcal{A}_j\}_{j=1}^{M}$ denote $M$ image transformations where $\mathcal{A}_1$ is the identity. The TTA-averaged prediction is
\begin{equation}
\bar{p}_{i+k}
    \;=\;
    \frac{1}{M}\sum_{j=1}^{M}
      g(\mathcal{A}_j(y_{i-n+1:i})),
    \label{eq:tta}
\end{equation}
where $g$ is the safety confidence predictor (Definition~\ref{def:predictor}). We use $M{=}9$: the original image plus eight augmentations with randomly sampled photometric parameters (contrast and saturation drawn uniformly from $[0.7, 1.3]$; Gaussian noise with $\sigma \sim \mathcal{U}(0, 0.05)$).


Intuitively, a correct prediction is stable under small appearance perturbations, so averaging preserves its confidence, whereas predictions near a decision boundary diverge under augmentation and average toward $0.5$~\cite{hekler2023tta}. The resulting $\bar{p}_{i+k}$ is the base prediction for temperature scaling, and TTA adds 12.9\,ms of overhead per sample (Table~\ref{tab:runtime} in the Appendix).
\subsubsection{Anomaly-Conditioned Temperature Scaling}
\label{sec:temperature_scaling}

We extend standard temperature scaling~\cite{guo2017calibration} by making the effective temperature depend on anomaly evidence:

\begin{equation}
    T_{\mathrm{eff}}
    \;=\;
    T_k \cdot \exp\bigl(w_\rho\,\rho_i + w_\delta\,\delta_i\bigr),
    \label{eq:teff}
\end{equation}
\begin{equation}
    \tilde{p}_{i+k}
    \;=\;
    \operatorname{sigmoid}\!\left(
      \frac{\operatorname{logit}(\bar{p}_{i+k})}{T_{\mathrm{eff}}}
    \right),
    \label{eq:calibrated}
\end{equation}
where $T_k > 0$ is a per-horizon base temperature, $w_\rho \geq 0$ and $w_\delta \geq 0$ are two scalar weights, and $\operatorname{sigmoid}(\cdot)$ denotes the logistic sigmoid function.


We train the calibrator \textit{jointly} across a set of prediction horizons $\mathcal{K} = \{1, 5, 10, \ldots, 50\}$. The calibrator has $|\mathcal{K}|+2$ learnable parameters: one base temperature $T_k$ per horizon $k \in \mathcal{K}$, plus two shared anomaly weights $w_\rho$ and $w_\delta$. These are fit in two steps on the withheld calibration set.
First, for each horizon $k$, we fit $T_k$ on in-distribution calibration data $\mathcal{D}_{\mathrm{in}}^{\mathrm{cal}}$ by minimizing negative log-likelihood (NLL) over TTA-averaged predictions, matching standard temperature scaling~\cite{guo2017calibration}.
Second, we fit the shared weights $w_\rho$ and $w_\delta$ on a mixture of $\mathcal{D}_{\mathrm{in}}^{\mathrm{cal}}$ and the augmented set $\mathcal{D}_{\mathrm{aug}}$~\cite{hendrycks2019oe} by minimizing NLL using L-BFGS-B with bound constraints $w_\rho, w_\delta \ge 0$.
When $\rho_i \approx \delta_i \approx 0$, the calibrator reduces to standard temperature scaling with $T_k$. As anomaly evidence grows, $T_{\mathrm{eff}}$ increases, pulling probabilities toward $0.5$ and reducing overconfidence.




\section{EXPERIMENTAL EVALUATION}
\label{sec:experimental_setup}

\subsection{Experimental Setup}
We evaluate the proposed framework using a physical DonkeyCar platform~\cite{viitala2021learning}. Autonomous racing is a compelling testbed: controllers must map camera images to actions under tight latency, and even brief miscalibration can cause track departures. The vehicle is equipped with an NVIDIA Jetson Nano for on-board inference and a front-facing camera. 
Images are resized to $64 \times 64 \times 3$ pixels for model input. 
Experiments use a custom indoor track (${\sim}20$\,m circumference, tile surface) with sharp turns.

The overall pipeline consists of a frozen world model $\omega$ for anomaly scoring and a separate image-based safety predictor $g$ for $k$-step safety confidence.
The world model encoder and decoder, $(e,d)$, is a VAE operating on $64{\times}64$ RGB images to $z_t\in\mathbb{R}^{64\times4\times4}$. The world model latent dynamics $\hat{f}_\theta$ is a two-layer ConvLSTM (128 hidden channels, dropout 0.3) predicting latent evolution over horizon $K_{\max}$.
The safety predictor $g$ is a CNN (4 conv $+$ 3 FC layers) outputting an estimated probability of safe/unsafe.

Both the world model (VAE reconstruction $+$ latent prediction loss, AdamW) and safety predictor (binary cross-entropy, AdamW) are trained on $\mathcal{D}_{\mathrm{in}}^{\mathrm{train}}$ and frozen thereafter; full hyperparameters are in Appendix~\ref{app:training_details}.


\looseness=-1
\noindent
\textbf{Anomaly Injection Protocols.}
\emph{(1) Visual Anomalies:} 
We introduce environmental and sensor-level corruptions inspired by the ImageNet-C benchmark~\cite{hendrycks2019robustness} at five severity levels. For real-world evaluation, we focus on two protocols: global luminance reduction to simulate low-light conditions (\textit{Dark}) and Gaussian blur to simulate defocused or degraded optics (\textit{Blur}). During augmented-data construction ($\mathcal{D}_{\mathrm{aug}}$), we apply eight corruption types that are disjoint from the test protocols: JPEG compression, desaturation, fog, speckle noise, salt-and-pepper noise, pixelation, color jitter, and frost. Notably, no brightness-, contrast-, or blur-family corruptions are included in $\mathcal{D}_{\mathrm{aug}}$, so the calibrator is not trained on the same corruption families as the test-time visual protocols (Dark/Blur). We emphasize that TTA applies mild photometric jitter only at inference time to smooth predictions, rather than to provide training-time corruption coverage.

\begin{figure}[t]
  \centering
  \includegraphics[width=1\linewidth]{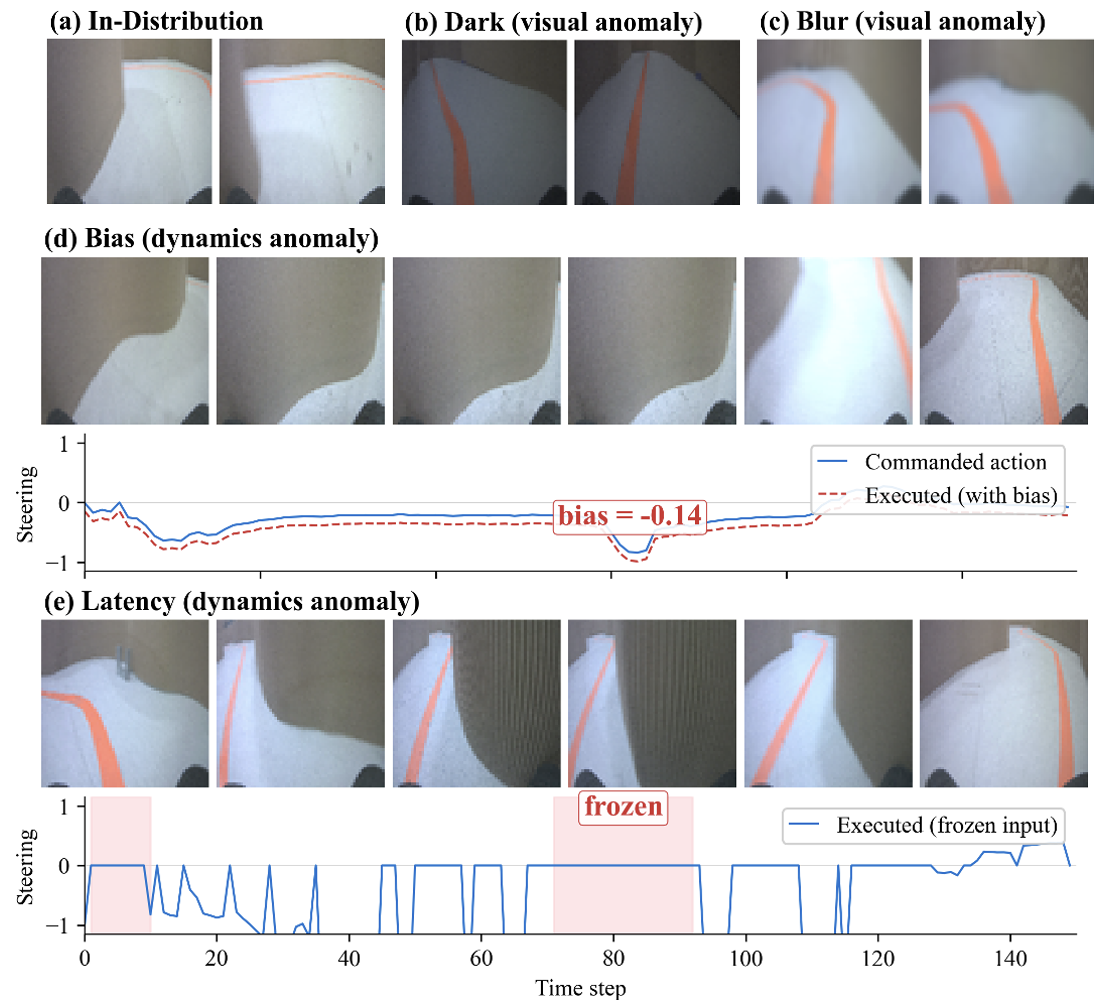}
    \vspace{-5mm}
  \caption{Images of in-distribution data and four anomalies on the DonkeyCar platform.
    Visual anomalies (b,\,c) are observable from raw frames,
    while dynamics anomalies (d,\,e) only manifest in steering signals
    with normal-looking images.}
  \label{fig:ood-examples}
  \vspace{-4mm}
\end{figure}

\looseness=-1
\emph{(2) Dynamics Anomalies:} 
To simulate hardware degradation or unmodeled physics, we manipulate the control and processing pipeline. For real-world evaluation, we define two protocols: \textit{Bias}, which injects additive Gaussian noise into the throttle and steering command streams, and \textit{Latency}, which introduces variable processing delays and frame freezing where the image feed remains static for a set duration before resuming. During augmented-data construction ($\mathcal{D}_{\mathrm{aug}}$), we apply three dynamics perturbation types that are disjoint from the test protocols: temporal speed mismatch (2$\times$ subsampling), cross-sequence action substitution, and within-sequence action offset. No action-noise, frame-freezing, or actuation-delay perturbations are included in $\mathcal{D}_{\mathrm{aug}}$, so the calibrator is not trained on perturbations from the same family as the test-time dynamics protocols (Bias/Latency). These two protocols target the dominant failure modes of low-cost robotic platforms. \textit{Bias} reproduces miscalibrated or worn actuators and uneven surface friction, which manifest as a persistent additive offset on the commanded steering and throttle --- as on hobby-grade servos and brushed motors whose response drifts with temperature, battery sag, and mechanical wear. \textit{Latency} reproduces compute contention and dropped frames on embedded inference hardware (here a Jetson Nano), where buffer overruns, thermal throttling, or sensor-readout stalls make the controller keep acting on stale or frozen observations. Both perturbations leave individual frames visually plausible, so they stress the dynamics channel rather than the perception channel.
Representative frames for in-distribution driving and the four real-world protocols (Dark, Blur, Bias, Latency) are shown in Fig.~\ref{fig:ood-examples}.

\noindent
\textbf{Data Acquisition and Training.}
We collect 25 driving sequences on the physical track under nominal conditions. Each sequence comprises 500--1{,}000 frames (average ${\sim}748$) of $64{\times}64$ RGB images, paired with steering/throttle actions and per-frame binary safety labels.


The nominal set is approximately balanced (52.5\% safe); OOD sets vary (Dark 63.4\%, Blur 70.7\%, Bias 47.6\%, Latency 42.6\%).
We evaluate horizons $k \in \mathcal{K}=\{1,5,10,\ldots,50\}$ with $K_{\max}=50$.

\begin{figure}[t]
  \centering
  \includegraphics[width=\linewidth]{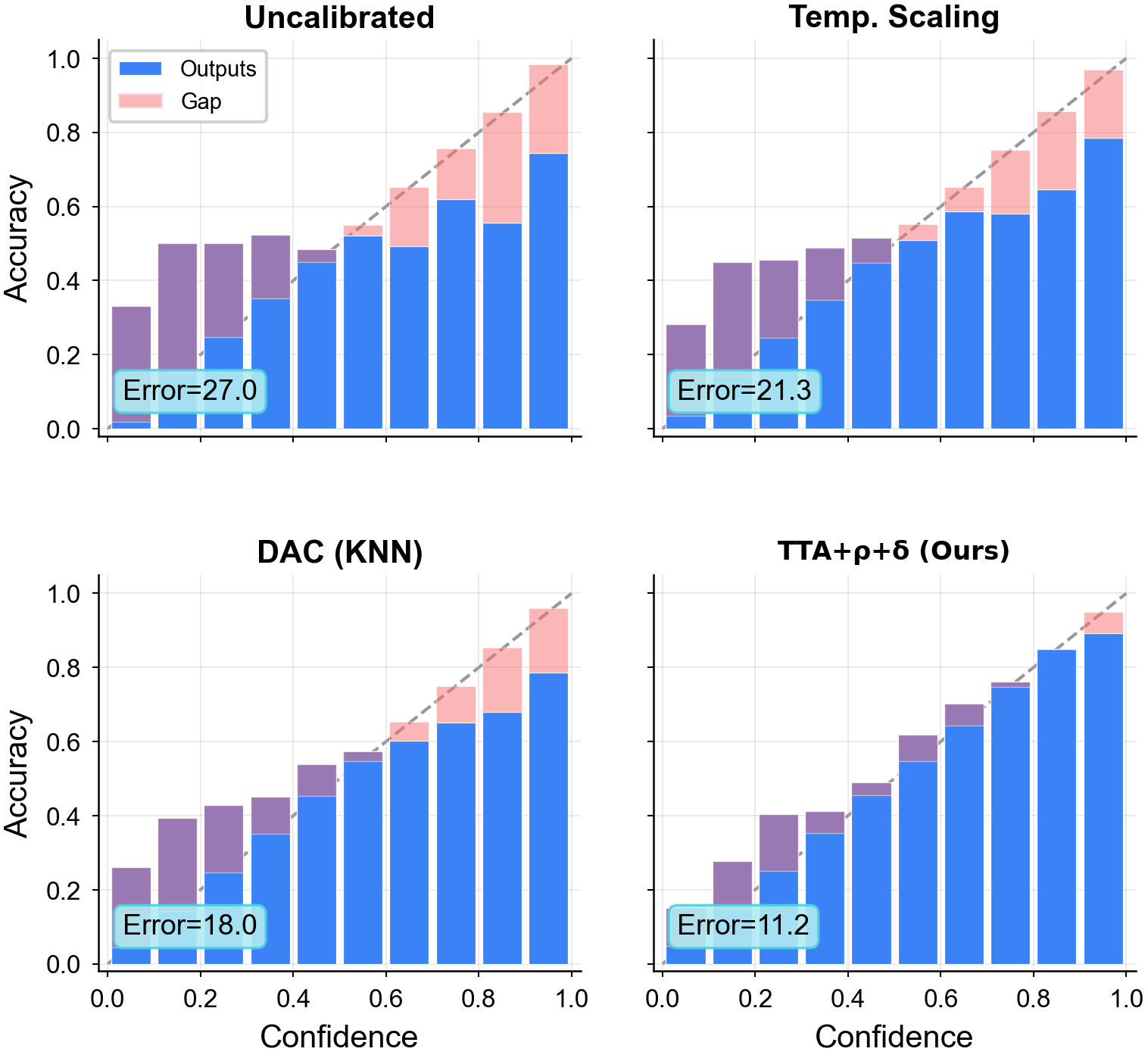}
  \caption{Reliability diagrams aggregated over all four OOD protocols and horizons $k \in \{5,\ldots,50\}$.
  Blue bars denote empirical accuracy per confidence bin; pink regions show the calibration gap.
  The dashed diagonal represents perfect calibration.
  Our method (bottom-right, Error$=$11.2) achieves the smallest gap across all confidence levels.}
  \label{fig:reliability}
  \vspace{-5mm}
\end{figure}
We split the nominal dataset $\mathcal{D}_{\mathrm{in}}$ into $\mathcal{D}_{\mathrm{in}}^{\mathrm{train}}$ (68\%), $\mathcal{D}_{\mathrm{in}}^{\mathrm{cal}}$ (16\%), and a held-out nominal test set $\mathcal{D}_{\mathrm{in}}^{\mathrm{test}}$ (16\%) at the level of whole driving sequences: consecutive frames within a sequence are nearly identical, so a frame-level split would place near-duplicates of test frames into the training set.
The world model and safety predictor are trained on $\mathcal{D}_{\mathrm{in}}^{\mathrm{train}}$ and remain frozen thereafter.

To train the anomaly-conditioned calibrator, we construct $\mathcal{D}_{\mathrm{aug}}$ from dynamics-augmented sequences (temporal speed mismatch, cross-sequence action substitution, and within-sequence action offset; perturbation types that overlap with the test protocols (action noise, frame freezing, actuation delay) are excluded) and visual-augmented sequences (ImageNet-C corruptions excluding the brightness, contrast, and blur families used in test protocols). The ID calibration data is obtained via 5-fold cross-validation on $\mathcal{D}_{\mathrm{in}}^{\mathrm{cal}}$ and pooled with $\mathcal{D}_{\mathrm{aug}}$, subsampled to approximate balance when the augmented portion exceeds twice the ID count. Per-horizon base temperatures $T_k$ are first fit on $\mathcal{D}_{\mathrm{in}}^{\mathrm{cal}}$ by minimizing NLL; the shared weights $w_\rho, w_\delta \geq 0$ are then fit on the balanced pool via L-BFGS-B. Calibration is evaluated on four real-world anomaly protocols never seen during training (Dark, Blur, Bias, Latency), totalling 22{,}253 frames; per-protocol details are provided in Appendix~\ref{app:per_protocol}.
The covariance regularization parameter for the Mahalanobis distance (Eq.~\ref{eq:delta}) is set to $\lambda=10^{-6}$.


\noindent\textbf{Safety labels.}
Labels are generated offline from an overhead camera by tracking a colored marker to compute cross-track error (CTE); a frame is unsafe if $\mathrm{CTE}_t>\tau_{\mathrm{cte}}$. While a geometric proxy rather than a direct collision indicator, CTE is the standard safety metric for small-scale racing platforms~\cite{viitala2021learning}. The overhead camera is not available at deployment.

\begin{table}[h]
\centering
\caption{OOD detection performance (AUROC $\uparrow$) of anomaly scores across four real-world anomaly protocols. ID test set ($n{=}890$) vs.\ each OOD type.}
\label{tab:ood_detection}
\renewcommand{\arraystretch}{1.15}
\setlength{\tabcolsep}{5pt}
\begin{tabular}{@{}lcccc|c@{}}
\toprule
\textbf{Method} & \textbf{Dark} & \textbf{Blur} & \textbf{Bias} & \textbf{Latency} & \textbf{Mean} \\
\midrule
\multicolumn{6}{@{}l}{\textit{Prediction-derived baselines}} \\
MSP~\cite{hendrycks2017baseline}            & 0.534 & 0.538 & 0.537 & 0.557 & 0.542 \\
Entropy                                      & 0.542 & 0.538 & 0.536 & 0.554 & 0.543 \\
MC Dropout~\cite{gal2016dropout}             & 0.461 & 0.453 & 0.490 & 0.535 & 0.485 \\
\midrule
\multicolumn{6}{@{}l}{\textit{Feature-based baseline}} \\
Feature Norm~\cite{yu2023featnorm}           & 0.648 & 0.689 & 0.562 & 0.732 & 0.658 \\
\midrule

$\rho$ (perception)            & \textbf{0.899} & \textbf{0.798} & 0.813 & 0.809 & 0.830 \\
$\delta$ (dynamics)            & 0.608 & 0.516 & 0.669 & 0.558 & 0.588 \\
Ours ($\rho{+}\delta$)         & 0.897 & 0.785 & \textbf{0.843} & \textbf{0.810} & \textbf{0.834} \\

\bottomrule
\end{tabular}
\end{table}

\noindent\textbf{Baselines.}
Our baselines are chosen to cover the established families of \emph{post-hoc} calibration --- the setting most comparable to ours, since none retrain the base predictor. We include the two classical parametric methods (Temperature~\cite{guo2017calibration} and Platt~\cite{platt1999probabilistic} Scaling), two classical non-parametric methods (Isotonic Regression~\cite{zadrozny2002transforming} and Histogram Binning~\cite{zadrozny2001obtaining}), and DAC~\cite{tomani2023beyond}, a recent density-aware method that, like ours, conditions temperature on a shift signal (KNN feature density) and is therefore the most direct competitor. This spans the parametric/non-parametric/density-aware axes while keeping every baseline in the retraining-free regime. We report horizon-averaged Expected Calibration Error ($\overline{\mathrm{ECE}}$), Adaptive Calibration Error (ACE), Static Calibration Error (SCE)~\cite{nixon2019measuring} and AUROC; ablations isolate TTA, $\rho$, and $\delta$. All definitions of the evaluation metrics can be found in Appendix~\ref{app:metrics}.
For OOD detection in Table~\ref{tab:ood_detection}, we additionally report standard confidence-derived heuristics (MSP~\cite{hendrycks2017baseline}, entropy, MC Dropout~\cite{gal2016dropout}) and a representation-distance baseline (Feature Norm~\cite{yu2023featnorm}) as sanity checks.

\subsection{Experimental Results}
\label{sec:results}

\looseness=-1
The evaluation revolves around three questions:
(Q1)~Can the proposed anomaly scores reliably distinguish in-distribution from OOD data?
(Q2)~Does anomaly-informed calibration outperform standard post-hoc calibrators under distribution shift?
(Q3)~What is the individual contribution of each component to anomaly-informed calibration?

All calibration results are reported as mean$\,\pm\,$std over 5 independent random splits of the ID calibration set.
Throughout, we report $\overline{\mathrm{ECE}}$, computed as a trapezoidal-rule mean over horizons $k \in \{5,10,\ldots,50\}$ and then averaged across the four OOD protocols.

\noindent
\textbf{Base Predictor Degradation.}
The base CNN predictor degrades substantially under shift: mean AUROC drops from $0.944$ (ID) to $0.676$/$0.774$/$0.763$/$0.756$ on Dark/Blur/Bias/Latency. Degradation increases with horizon; for Dark, AUROC at $k{=}50$ falls to $0.547$ (Table~\ref{tab:safety_auroc}).

\begin{table}[t]
\centering
\caption{Safety classification AUROC of the base CNN predictor.}
\label{tab:safety_auroc}
\setlength{\tabcolsep}{5pt}
\begin{tabular}{lcccc|c}
\toprule
 & \multicolumn{4}{c}{Prediction horizon $k$} & \\
\cmidrule(lr){2-5}
Condition & 5 & 15 & 30 & 50 & Mean \\
\midrule
ID (normal)        & 0.986 & 0.966 & 0.898 & 0.897 & 0.944 \\
\midrule
Dark (visual)      & 0.834 & 0.797 & 0.612 & 0.547 & 0.676 \\
Blur (visual)      & 0.936 & 0.867 & 0.712 & 0.682 & 0.774 \\
Bias (dynamics)    & 0.892 & 0.808 & 0.723 & 0.710 & 0.763 \\
Latency (dynamics) & 0.871 & 0.822 & 0.689 & 0.705 & 0.756 \\
\bottomrule
\end{tabular}
\vspace{-3mm}
\end{table}

\noindent
\textbf{OOD Detection Quality (Q1).}
Table~\ref{tab:ood_detection} reports ID-vs-OOD AUROC.
Prediction-derived baselines (MSP, Entropy, MC Dropout) are near chance (mean AUROC $\leq 0.543$), confirming that the base predictor's confidence alone cannot distinguish shifted inputs.
Our perception score $\rho$ alone already achieves 0.830 mean AUROC (Dark: 0.899; Blur: 0.798), outperforming all baselines including Feature Norm (0.658). Fusing the dynamics score ($\rho{+}\delta$) raises the mean to 0.834 and is strongest on the dynamics-Bias protocol (0.843), while $\rho$ remains best on the visual protocols. Notably, $\rho$ already detects dynamics anomalies: in closed loop, a dynamics fault degrades the trajectory and renders the observation visually OOD. Synthetic augmentations transfer to real-world OOD detection across all four anomaly types and severity levels, consistent with Assumption~\ref{assump:transfer}. We note that Assumption~\ref{assump:transfer} is a sufficient condition for calibrator generalization, not a quantity we claim to verify exactly; the consistent improvements across four qualitatively distinct anomaly types at multiple severity levels provide indirect support for both monotonicity and approximate sufficiency.
This answers Q1: the world-model-derived scores substantially outperform all simple baselines in OOD discrimination.


\newcommand{\std}[1]{\ensuremath{{\scriptscriptstyle\pm #1}}}

\begin{table}[t]
\centering
\caption{Calibration performance: $\overline{\mathrm{ECE}}$ ($\downarrow$), mean$\,\pm\,$std over 5 splits. Best in \textbf{bold}.}
\label{tab:baseline_ece}
\renewcommand{\arraystretch}{1.15}
\setlength{\tabcolsep}{3pt}
\scriptsize
\resizebox{\columnwidth}{!}{%
\begin{tabular}{@{}l cccc|c@{}}
\toprule
\textbf{Method} & \textbf{Dark} & \textbf{Blur} & \textbf{Bias} & \textbf{Lat.} & \textbf{Mean} \\
\midrule
Uncalibrated
& .365\std{0.000} & .221\std{0.000} & .251\std{0.000} & .259\std{0.000} & .274\std{0.000} \\
Temp.\ Scaling
& .305\std{0.003} & .166\std{0.003} & .194\std{0.003} & .206\std{0.004} & .218\std{0.003} \\
Platt
& .311\std{0.004} & .172\std{0.005} & .192\std{0.004} & .205\std{0.007} & .220\std{0.004} \\
Isotonic
& .332\std{0.006} & .189\std{0.005} & .202\std{0.005} & .208\std{0.003} & .233\std{0.004} \\
Hist.\ Binning
& .345\std{0.005} & .207\std{0.004} & .225\std{0.003} & .238\std{0.005} & .254\std{0.004} \\
DAC~\cite{tomani2023beyond}
& .266\std{0.006} & .136\std{0.006} & .152\std{0.003} & .181\std{0.005} & .184\std{0.004} \\
\midrule
Ours (TTA+$\rho$+$\delta$)
& \textbf{.165}\std{0.007} & \textbf{.098}\std{0.003} & \textbf{.076}\std{0.005} & \textbf{.127}\std{0.005} & \textbf{.116}\std{0.003} \\
\bottomrule
\end{tabular}
}
\end{table}


\noindent
\textbf{Calibration Performance Against Baselines (Q2).}
Table~\ref{tab:baseline_ece} (horizon-averaged) and Fig.~\ref{fig:baseline_ece} (per horizon $k$) show that standard post-hoc calibrators reduce ECE but degrade under shift, whereas our method (TTA+$\rho$+$\delta$) attains the lowest error at every horizon and protocol --- including the long horizons where the base predictor is most degraded (Table~\ref{tab:safety_auroc}).
DAC, the strongest baseline (mean ECE $0.184$), confirms that feature-space density helps; our method further reduces ECE to $0.116$ (37\% improvement). On the held-out nominal test set $\mathcal{D}^{\mathrm{test}}_{\mathrm{in}}$, our calibrator preserves in-distribution performance, achieving a mean ECE of $0.121$.
These rankings are consistent under ACE and SCE~\cite{nixon2019measuring} (Fig.~\ref{fig:ace_sce}). The reliability diagrams in Fig.~\ref{fig:reliability} confirm that our method nearly eliminates overconfidence, reducing aggregated error from $27.0$ to $11.2$.
All improvements over baselines are statistically significant (Wilcoxon signed-rank, $N{=}20$ paired observations across 4 OOD protocols $\times$ 5 seeds): $p<10^{-4}$ vs.\ Temperature Scaling, DAC, TTA, and TTA+$\delta$. Only TTA+$\rho$ vs.\ our full model is non-significant ($p{=}0.66$), indicating $\delta$'s contribution is complementary rather than a mean-ECE reduction.



\begin{table}[t]
\centering
\caption{Ablation: $\overline{\mathrm{ECE}}$ ($\downarrow$), mean$\,\pm\,$std over 5 splits.}
\label{tab:ablation}
\renewcommand{\arraystretch}{1.15}
\setlength{\tabcolsep}{3pt}
\scriptsize
\resizebox{\columnwidth}{!}{%
\begin{tabular}{@{}l cccc|c@{}}
\toprule
\textbf{Method} & \textbf{Dark} & \textbf{Blur} & \textbf{Bias} & \textbf{Lat.} & \textbf{Mean} \\
\midrule
Shrinkage ($\rho{+}\delta$)      
& .261\std{0.015} & .126\std{0.007} & .143\std{0.014} & .166\std{0.009} & .174\std{0.010} \\
Logit ($\rho{+}\delta$, w/o TTA) 
& .195\std{0.012} & .109\std{0.008} & .104\std{0.010} & .160\std{0.008} & .142\std{0.009} \\
\midrule
TTA                                
& .253\std{0.005} & .114\std{0.003} & .121\std{0.006} & .162\std{0.007} & .162\std{0.005} \\
TTA+$\delta$                        
& .225\std{0.009} & .110\std{0.003} & .097\std{0.008} & .159\std{0.009} & .148\std{0.007} \\
TTA+$\rho$                       
& \textbf{.162}\std{0.011} & \textbf{.094}\std{0.006} & .078\std{0.004} & .128\std{0.004} & .116\std{0.005} \\
TTA+$\rho{+}\delta$ (Ours)       
& .165\std{0.007} & .098\std{0.003} & \textbf{.076}\std{0.005} & \textbf{.127}\std{0.005} & \textbf{.116}\std{0.003} \\
\bottomrule
\end{tabular}
}
\vspace{-3mm}
\end{table}



\noindent
\textbf{Ablation Study (Q3).}
Table~\ref{tab:ablation} isolates each component (all variants share the same frozen predictor).
\noindent\textit{The perception score provides the dominant calibration signal.}
Adding $\rho$ on top of TTA reduces OOD-average ECE from $0.162$ to $0.116$ (28\%), with the largest gains on Dark and Bias.

\noindent\textbf{Perception-dynamics gap is not significant in a vision control loop.}
Adding $\delta$ on top of TTA+$\rho$ does not significantly change the mean ECE ($p{=}0.66$): in closed-loop deployment, a dynamics fault drives the vehicle off-track, so the resulting observations are themselves off-manifold and already captured by $\rho$; the perception-dynamics gap is therefore small in this closed-loop testbed. Still, $\delta$ provides a complementary, interpretable channel that responds directly to actuation perturbations, raising OOD-detection AUROC on the Bias protocol from $0.813$ to $0.843$ (Table~\ref{tab:ood_detection}).

\noindent\textbf{Logit-space scaling and TTA-free operation.}
At matched inputs ($\rho{+}\delta$, no TTA), logit-space scaling outperforms shrinkage ($0.142$ vs.\ $0.174$, 18\%) and even TTA alone ($0.162$), showing that anomaly conditioning can substitute for TTA in latency-constrained settings.

\section{DISCUSSION AND CONCLUSION}

We presented an anomaly-informed calibration framework that fuses TTA with world-model-derived perceptual and dynamics anomaly scores, reducing OOD-average ECE from $0.184$ to $0.116$ (37\% over the best baseline) on a physical DonkeyCar under four unseen anomaly protocols without retraining.
Our central finding is that the perception-dynamics gap is insignificant in a tightly coupled perception-control loop: a dynamics fault degrades the trajectory until the observation is itself OOD, so the perceptual score $\rho$ carries the calibration gains (28\% ECE reduction, TTA handling visual robustness implicitly) while the disentangled dynamics score $\delta$ adds no significant calibration ($p{=}0.66$). We therefore present $\delta$ not as a co-equal contribution but as an honest, mostly negative result: it still improves OOD detection on the pure actuation-bias protocol ($0.813{\to}0.843$ AUROC) and is the principled signal for settings (e.g., open-loop control or faults that do not propagate to vision) where the gap persists.
Transfer from synthetic augmentations to real-world protocols supports Assumption~\ref{assump:transfer}, though extrapolation beyond the training severity range remains a limitation.
Future work includes closed-loop safety integration, online calibrator updates, and conformal prediction~\cite{vovk2005algorithmic} for finite-sample calibration guarantees.
A further direction is to incorporate language-model-based signals. Vision-language models can supply a \emph{semantic} anomaly score that flags high-level, on-manifold shifts (e.g., unfamiliar scene layouts or out-of-context objects) that pixel-level reconstruction error misses~\cite{mao2024language}, providing a third channel alongside our perceptual ($\rho$) and dynamics ($\delta$) scores; building on our prior work that quantifies and calibrates confidence over the temporal structure of language-model reasoning~\cite{mao2025temporalizing,mao2026rcc,mao2026confidence}, a language model could further fuse these heterogeneous scores into human-readable explanations of why confidence was lowered, aiding operator oversight.
Moreover, as vision-language-action (VLA) policies~\cite{zitkovich2023rt2,kim2024openvla,black2024pi0} are increasingly deployed as end-to-end controllers, our calibration framework, which is agnostic to the underlying policy, could be extended to calibrate their safety confidence, where language-conditioned action introduces new dynamics-anomaly modes such as instruction-action mismatch that the dynamics score~$\delta$ is naturally positioned to capture.
\bibliographystyle{IEEEtran}
\bibliography{ref}
\clearpage



\appendix
\appsection{Training and Implementation Details}
\label{app:training_details}

Tables~\ref{tab:training_hyper} and~\ref{tab:calib_protocol} together list all training, calibration, and baseline hyperparameters.
The world model consists of a convolutional VAE operating on $64{\times}64$ images and a ConvLSTM predictor that autoregressively rolls out latent states given a context of $m{=}100$ frames.
Both are trained with AdamW and cosine-annealed learning rates.
The CNN safety classifier is trained directly on $64{\times}64$ RGB images with a binary cross-entropy objective.

\noindent\textbf{LSTM predictor details and rationale.}
The latent dynamics predictor $\hat f_\theta$ is a two-layer convolutional LSTM operating directly on the VAE latent tensor $z_t\in\mathbb{R}^{64\times4\times4}$, with $128$ hidden channels and $3{\times}3$ convolutional gates normalized by GroupNorm.
The two-dimensional control action (steering, throttle) is embedded by a small convolutional encoder into a $4{\times}4$ map and concatenated with the latent input at each step.
Instead of predicting the next latent directly, the network predicts a residual $\Delta z_t$ combined through a learned per-element gate, $z_{t+1}=z_t+\mathrm{gate}\odot\Delta z_t$, which stabilizes long-horizon rollouts.
Two dropout mechanisms are used and \emph{both remain active at test time} to produce the MC Dropout uncertainty features of $\delta$: an inter-layer $\mathrm{Dropout2d}$ ($p{=}0.3$) and a variational recurrent dropout on the hidden state ($p{=}0.15$).
Beyond the optimizer settings in Table~\ref{tab:training_hyper}, training uses $100$-frame context/$100$-frame target windows and, to keep the autoregressive rollout accurate at the $K_{\max}{=}50$ scoring horizon, \emph{scheduled sampling} (annealing the teacher-forcing probability from $1$ to $0$ over the first $40$ epochs) and an \emph{open-loop rollout curriculum} that progressively lengthens the free-running horizon.
The two-layer, $128$-channel width keeps the predictor lightweight ($2.4$M parameters, Table~\ref{tab:runtime}) and well-matched to the low-dimensional $4{\times}4{\times}64$ latent; the dropout rates are set high enough to yield informative MC-dropout uncertainty while preserving deterministic rollout accuracy.

For post-hoc calibration, per-horizon temperatures $T_k$ are obtained by minimizing negative log-likelihood (NLL) via bounded scalar optimization on the calibration split.
The anomaly-conditioned calibrator weights $(w_\rho, w_\delta)$ are fit on the pooled calibration set $\mathcal{D}^{\mathrm{cal}}_{\mathrm{in}} \cup \mathcal{D}_{\mathrm{aug}}$ (5-fold CV for robustness), with all real OOD data held out for evaluation.
The Mahalanobis distance for the dynamics score uses a covariance regularization of $\lambda{=}10^{-6}$ with StandardScaler pre-processing.
We verified that the dynamics score is insensitive to this choice: varying $\lambda \in \{10^{-8}, 10^{-6}, 10^{-4}, 10^{-2}\}$ changes the OOD-average ECE by less than $0.0002$ ($0.1164$ at $\lambda{=}10^{-6}$) and the dynamics-score AUROC by less than $0.001$, since $\lambda$ only stabilizes the covariance inverse rather than affecting the score.
Such small diagonal loading is standard practice for numerically stabilizing Mahalanobis-based OOD scores~\cite{lee2018simple}, with values in $[10^{-6},10^{-3}]$ commonly used.

\begin{table}[h]
\centering
\caption{Training hyperparameters for each pipeline component.}
\label{tab:training_hyper}
\small
\setlength{\tabcolsep}{3pt}
\begin{tabular}{@{}l>{\raggedright\arraybackslash}p{5.5cm}@{}}
\toprule
\textbf{Component} & \textbf{Hyperparameters} \\
\midrule
VAE ($64{\times}64$) &
  AdamW, lr $=10^{-3}$, weight decay $10^{-5}$, batch 32, 50 epochs, latent tensor $4{\times}4{\times}64$ (64 channels) \\
LSTM predictor &
  AdamW, lr $=10^{-4}$, weight decay $10^{-4}$, batch 64, 60 epochs, hidden 128 ($2$ layers), context $m{=}100$, dropout $0.3$ / recurrent $0.15$, cosine annealing ($\eta_{\min}{=}10^{-6}$) \\
CNN classifier &
  AdamW, lr $=10^{-3}$, weight decay $10^{-5}$, batch 128, 50 epochs \\
\midrule
Temperature scaling &
  Per-horizon $T_k$ via bounded scalar minimization of NLL \\
Anomaly calibrator &
  L-BFGS-B, $w_\rho, w_\delta \geq 0$, 5-fold CV on $\mathcal{D}^{\mathrm{cal}}_{\mathrm{in}} \cup \mathcal{D}_{\mathrm{aug}}$ \\
Mahalanobis $\delta$ &
  $\lambda = 10^{-6}$, StandardScaler pre-processing \\
\bottomrule
\end{tabular}
\end{table}

\bigskip
\appsection{Evaluation Metrics}
\label{app:metrics}

\noindent We evaluate calibration quality with three complementary errors --- Expected, Adaptive, and Static Calibration Error (ECE, ACE, SCE) --- and discriminative ability with AUROC.

\begin{itemize}
    \item \emph{Expected Calibration Error (ECE)~\cite{guo2017calibration}.}
We report ECE with $B$ bins:
\end{itemize}
\begin{equation}
\mathrm{ECE}=\sum_{b=1}^{B}\frac{|\mathcal{I}_{b}|}{N}
    \left|\mathrm{acc}(\mathcal{I}_{b})-\mathrm{conf}(\mathcal{I}_{b})\right|,
\end{equation}
where $\mathrm{acc}(\mathcal{I}_{b})$ and $\mathrm{conf}(\mathcal{I}_{b})$ denote the empirical accuracy and mean predicted confidence in bin $\mathcal{I}_b$. Lower ECE indicates better calibration; ECE uses $B{=}15$ equal-width bins on $[0,1]$.

\begin{itemize}
    \item \emph{Adaptive Calibration Error (ACE)~\cite{nixon2019measuring}.}
ACE replaces the equal-width bins of ECE with $B$ \emph{equal-mass} bins (samples sorted by confidence, $|\mathcal{I}_b|{=}N/B$) and averages the per-bin gap uniformly:
\end{itemize}
\begin{equation}
\mathrm{ACE}=\frac{1}{B}\sum_{b=1}^{B}
    \left|\mathrm{acc}(\mathcal{I}_{b})-\mathrm{conf}(\mathcal{I}_{b})\right|,
\end{equation}
Equal-mass binning prevents sparsely populated high-confidence bins from dominating the estimate.

\begin{itemize}
    \item \emph{Static Calibration Error (SCE)~\cite{nixon2019measuring}.}
SCE assesses calibration separately for each class and averages over the $C$ classes, using equal-width bins:
\end{itemize}
\begin{equation}
\mathrm{SCE}=\frac{1}{C}\sum_{c=1}^{C}\sum_{b=1}^{B}\frac{|\mathcal{I}_{b,c}|}{N}
    \left|\mathrm{acc}_{c}(\mathcal{I}_{b,c})-\mathrm{conf}_{c}(\mathcal{I}_{b,c})\right|,
\end{equation}
where $\mathrm{conf}_{c}$ is the predicted probability of class $c$ and $\mathrm{acc}_{c}$ its empirical frequency in the bin; we use $C{=}2$ classes and $B{=}15$ bins. Lower ACE and SCE both indicate better calibration.

\begin{itemize}
    \item \emph{Area Under the Receiver Operating Characteristic Curve (AUROC)}~\cite{fawcett2006roc}. We report AUROC for safety confidence scores and OOD detection scores:
\end{itemize}
\begin{equation}
    \text{AUROC} = \int_0^1 \text{TPR}\bigl(\text{FPR}^{-1}(u)\bigr)\, du,
\end{equation}
where TPR is the true positive rate, FPR is the false positive rate, and $u$ is the threshold. A value of $1.0$ means perfect discrimination, and $0.5$ means coin-flip performance.

\bigskip
\appsection{Dynamics Feature Vector}
\label{app:dynamics_features}

The dynamics feature vector $\mathbf{f}_i^{K}$ (overviewed in Sec.~\ref{sec:approach}) is a bank of generic descriptors of the control and latent streams, computed causally over a context window of $m_w{=}100$ steps.
The score $\delta_i$ is their Mahalanobis distance from the in-distribution manifold --- an unsupervised novelty score that rises with any deviation from nominal driving, regardless of its cause.
Intuitively, they summarize three things: how \emph{uncertain} the dynamics model is about its own prediction, how \emph{smooth and active} the steering command is, and how \emph{tightly the steering couples to the resulting latent (perceptual) change}.
They are built from three causal signals: (i)~the normalized \emph{steering command} $u_t$ sent to the vehicle (first difference $\Delta u_t=u_{t+1}-u_t$); (ii)~a per-step model-uncertainty signal $\sigma_t$, obtained by rolling out the latent dynamics $\hat f_\theta$ with dropout enabled $S{=}20$ times and taking, at each step, the standard deviation of the predicted latent across the $S$ samples (averaged over latent dimensions), so $\sigma_t$ is large when the model disagrees with itself; and (iii)~the latent magnitude $\ell_t=\lVert z_t\rVert_2$.
A step is \emph{held} when the steering is essentially constant ($|\Delta u_t|<0.02$) and \emph{actively turning} when the steering is well off-centre ($|u_t|>0.15$).
Features marked $^\dagger$ depend on the horizon $K$; all are scalars computed from the context and a single $\hat f_\theta$ forward pass (no future observations), so $\delta_i$ is available online at time $i$.

\begin{table}[h]
\centering
\caption{Definitions of the 12 generic dynamics descriptors $\mathbf{f}_i^{K}$.}
\label{tab:dynamics_features}
\footnotesize
\setlength{\tabcolsep}{3pt}
\renewcommand{\arraystretch}{1.25}
\begin{tabular}{@{}l>{\raggedright\arraybackslash}p{5.2cm}@{}}
\toprule
\textbf{Descriptor} & \textbf{Definition} \\
\midrule
\multicolumn{2}{@{}l}{\textit{MC Dropout uncertainty (3)}}\\
rollout-mean$^\dagger$ & mean of $\sigma_t$ over the $K$-step rollout \\
rollout-max$^\dagger$  & peak $\sigma_t$ over the rollout \\
volatility$^\dagger$   & standard deviation of $\sigma_t$ \\
\midrule
\multicolumn{2}{@{}l}{\textit{Action patterns (6)}}\\
longest steering hold       & length of the longest constant-steering run ($|\Delta u_t|<0.02$) \\
hold while turning          & the same run, counted only while actively turning ($|u_t|>0.15$) \\
held fraction while turning & fraction of actively-turning steps with constant steering ($|u_t|>0.15$) \\
reversal rate               & how often the steering flips direction (share of successive steering changes with opposite sign) \\
jerk                        & mean $|u_{t+1}{-}2u_t{+}u_{t-1}|$ (steering smoothness) \\
mean steering               & average steering command \\
\midrule
\multicolumn{2}{@{}l}{\textit{Temporal correlations (3)}}\\
steering--latent xcorr & peak normalized cross-correlation between steering changes $\Delta u_t$ and latent changes $\Delta\ell_t$ (lags $0$--$10$) \\
HF spectral ratio      & share of the steering signal's Fourier energy in its upper half (steering jitter) \\
response lag           & frame lag at which the steering--latent correlation peaks \\
\bottomrule
\end{tabular}
\end{table}

\bigskip
\appsection{Per-Protocol Statistics}
\label{app:per_protocol}

Table~\ref{tab:per_protocol} summarizes the four real-world OOD evaluation
protocols. All anomalies are injected before the controller during closed-loop
deployment on the physical track, so perturbations propagate through both the
input to the safety predictor and the resulting trajectory. None of these
sequences are seen during world-model, safety-predictor, or calibrator training.

\begin{table}[h]
\centering
\caption{Statistics of the four real-world OOD evaluation protocols.}
\label{tab:per_protocol}
\small
\setlength{\tabcolsep}{4pt}
\begin{tabular}{@{}l l c c c@{}}
\toprule
\textbf{Protocol} & \textbf{Type} & \textbf{Seqs.} & \textbf{Frames} & \textbf{\% Safe} \\
\midrule
Dark    & Visual    & 7  & 4{,}858  & 63.4 \\
Blur    & Visual    & 8  & 4{,}703  & 70.7 \\
Bias    & Dynamics  & 17 & 10{,}117 & 47.6 \\
Latency & Dynamics  & 5  & 2{,}575  & 42.6 \\
\midrule
\textbf{Total} & --- & 37 & 22{,}253 & --- \\
\bottomrule
\end{tabular}
\end{table}

\bigskip
\appsection{Computational Cost}
\label{app:runtime}

Table~\ref{tab:runtime} reports per-sample inference time for each pipeline component. All models use FP32 precision; batch size is 1 unless noted.

\begin{table}[h]
\centering
\caption{Per-sample inference time breakdown. MC Dropout samples are batched ($B{=}3$) for GPU parallelism. Model sizes: VAE 5.5M, LSTM 2.4M, CNN 2.6M parameters.}
\label{tab:runtime}
\small
\setlength{\tabcolsep}{4pt}
\begin{tabular}{@{}llr@{}}
\toprule
\textbf{Stage} & \textbf{Component} & \textbf{Time (ms)} \\
\midrule
\multirow{2}{*}{Perception}
  & VAE encode ($m{=}100$ frames)           & 2.7 \\
  & VAE decode (1 frame)                     & 2.2 \\
\midrule
\multirow{2}{*}{Dynamics}
  & LSTM deterministic (150 steps)           & 173.7 \\
  & MC Dropout ($B{=}3$, 150 steps)          & 279.4 \\
\midrule
\multirow{2}{*}{Safety pred.}
  & CNN inference (1 image)                  & 1.5 \\
  & TTA ($1{+}8$ augmented views)            & 14.4 \\
\midrule
\multirow{2}{*}{Anomaly scores}
  & $\rho$: VAE reconstruction (100 fr)      & 52.4 \\
  & $\delta$: Mahalanobis (12-d)             & $<$0.01 \\
\midrule
\multirow{3}{*}{Calibration}
  & Temperature Scaling                      & $<$0.01 \\
  & Ours (anomaly-conditioned TS)            & $<$0.01 \\
  & DAC (KNN, $k{=}10$)                      & 19.5 \\
\midrule[\heavyrulewidth]
\multicolumn{2}{@{}l}{\textbf{Pipeline totals}} & \\
\midrule
\multicolumn{2}{@{}l}{Base (world model + CNN + Temp Scaling)}    & 180.0 \\
\multicolumn{2}{@{}l}{+TTA}                                       & 192.9 \\
\multicolumn{2}{@{}l}{Full (ours: +TTA +MC +$\rho$ +$\delta$)}   & 524.7 \\
\bottomrule
\end{tabular}
\end{table}

The dominant cost is the sequential ConvLSTM rollout: the deterministic pass (100-step priming $+$ 50-step prediction) takes 173.7\,ms, and the batched MC Dropout pass adds 279.4\,ms.
The calibration step itself is negligible ($<$0.01\,ms), confirming that the overhead of anomaly-informed calibration over standard temperature scaling is due entirely to the anomaly score extraction, not the calibration function.
Compared to the Base pipeline, our full method adds 344.7\,ms (+1.9$\times$), of which 80\% is MC Dropout.

\paragraph{Online deployment}
The times above reflect \emph{cold-start} evaluation, where each sample is processed independently from a full context window.
In a streaming setting, three properties reduce the effective per-frame cost substantially:
\begin{enumerate}
  \item \textbf{Incremental hidden state.}
    The LSTM maintains a running hidden state across frames.
    Each new observation requires only a single forward step (${\approx}\,1.2$\,ms) rather than re-priming over the entire context.
  \item \textbf{Asynchronous anomaly scoring.}
    The anomaly scores $\rho_i$ and $\delta_i$ characterize sustained distributional shift and need not be recomputed every frame.
    Updating them every $N$ frames (e.g., $N{=}10$ at 10\,Hz) amortizes the MC Dropout and VAE reconstruction cost to ${\approx}\,33$\,ms/frame, running on a separate CUDA stream without blocking the control path.
  \item \textbf{Decoupled control and monitoring.}
    The safety prediction (CNN + TTA) runs on the critical control path at ${\approx}\,15$\,ms/frame (${\approx}\,67$\,Hz).
    Anomaly scoring and calibration operate asynchronously as a monitoring layer; the latest available score $\delta_i$ is used.
\end{enumerate}
Under this design, the real-time bottleneck is the CNN + TTA pass (${\approx}\,15$\,ms), which comfortably meets typical camera rates (10--30\,Hz).

\bigskip
\appsection{Calibration Protocol}
\label{app:calibration_protocol}

\begin{table}[h]
\centering
\caption{Calibration metrics and baseline hyperparameters.}
\label{tab:calib_protocol}
\small
\setlength{\tabcolsep}{3pt}
\begin{tabular}{@{}l p{5.2cm}@{}}
\toprule
\textbf{Item} & \textbf{Setting} \\
\midrule
ECE / ACE / SCE & $B{=}15$ bins \\
ECE binning & 15 equal-width bins on $[0,1]$ \\
ACE binning & 15 equal-mass bins (sorted by confidence) \\
SCE binning & 15 equal-width bins, averaged over classes \\
\midrule
Temp. Scaling & $T_k\in[0.1,20]$, per-horizon, NLL objective \\
Platt Scaling & $(a_k,b_k)$ per-horizon, L-BFGS-B on $\sigma(a_k\mathrm{logit}(p)+b_k)$ \\
Isotonic & \texttt{sklearn} IsotonicRegression, $y_{\min}{=}0.01$, $y_{\max}{=}0.99$, clip \\
Hist. Binning & 15 equal-width bins \\
\midrule
DAC features & CNN penultimate layer (fc2 ReLU), 128-D \\
DAC kNN & $k{=}10$, Euclidean distance \\
DAC norm. & mean $k$NN distance, normalized by $\mu{+}2\sigma$ and clipped to $[0,1]$ \\
DAC modulation & $T_{\mathrm{eff}} = T_k\cdot\exp(2\,d_n)$ \\
\bottomrule
\end{tabular}
\end{table}

\addtolength{\textheight}{-1cm}   




\clearpage

\clearpage

\end{document}